\documentclass[runningheads]{llncs}

\usepackage[T1]{fontenc}
\usepackage{graphicx}
\usepackage[hidelinks]{hyperref}
\usepackage{booktabs}
\usepackage{amsfonts}
\usepackage{amsmath}
\usepackage{multirow}
\usepackage{bbm}

\begin{document}

\title{Towards plausibility in time series counterfactual explanations}

\titlerunning{Towards plausibility in time series counterfactual explanations}

\author{Marcin Kostrzewa\inst{1}\orcidID{0009-0001-1146-9002} \and Krzysztof Galus\inst{1}\orcidID{0009-0008-3342-6477} \and
Maciej Zięba\inst{1, 2}\orcidID{0000-0003-4217-7712}}

\authorrunning{M. Kostrzewa, K. Galus, and M. Zięba}

\institute{Wrocław University of Science and Technology, Wrocław 50-347, Poland \\
\and Tooploox, Wrocław 53-601, Poland \\
\email{marcin.kostrzewa@pwr.edu.pl}}
\maketitle

\begin{abstract}

We present a new method for generating plausible counterfactual explanations for time series classification problems. The approach performs gradient-based optimization directly in the input space. To enforce plausibility, we integrate soft-DTW (dynamic time warping) alignment with $k$-nearest neighbors from the target class, which effectively encourages the generated counterfactuals to adopt a realistic temporal structure. The overall optimization objective is a multi-faceted loss function that balances key counterfactual properties. It incorporates losses for validity, sparsity, and proximity, alongside the novel soft-DTW-based plausibility component. We conduct an evaluation of our method against several strong reference approaches, measuring the key properties of the generated counterfactuals across multiple dimensions. The results demonstrate that our method achieves competitive performance in validity while significantly outperforming existing approaches in distributional alignment with the target class, indicating superior temporal realism. Furthermore, a qualitative analysis highlights the critical limitations of existing methods in preserving realistic temporal structure. This work shows that the proposed method consistently generates counterfactual explanations for time series classifiers that are not only valid but also highly plausible and consistent with temporal patterns.

\end{abstract}

\section{Introduction}

Explainable artificial intelligence has become essential as complex machine learning systems are increasingly deployed in high-stakes domains~\cite{ALI2023survey,LONGO2024xai,Yang2023survey}. Among various explainability methods, counterfactual explanations (CFEs) provide actionable insights by answering the question: \textit{what minimal changes to the input would alter the model’s prediction?}~\cite{guidotti2024counterfactual,Verma2024survey,Wachter2017CounterfactualEW}. Unlike feature-importance approaches, which merely highlight influential variables, CFEs specify concrete modifications required to achieve a desired outcome. Although significant progress has been made, most existing CFE methods have been developed for tabular data~\cite{guidotti2024counterfactual}.

This gap is, for instance, observed in time series domains. Time series classifiers are widely used in healthcare for tasks such as ECG analysis and patient monitoring \cite{Ullah2020ClassificationOA,Xu2019TowardsEE}, in finance for fraud detection and risk assessment \cite{Jurgovsky2018SequenceCFA}, and in industrial systems for anomaly detection and predictive maintenance \cite{Markovic2023TimeseriesADA}. Advances in time series classification have produced highly accurate architectures such as ROCKET \cite{Dempster2020Rocket}, InceptionTime \cite{Fawaz2019InceptionTime}, and Hive-Cote \cite{Middlehurst2021HiveCote}, yet their complexity makes their decision processes difficult to interpret. To ensure trust, it is crucial to understand which temporal patterns drive these decisions. CFEs offer an intuitive way to reason about such predictions by identifying how specific temporal regions would need to change to obtain a different outcome. However, for time series data, the utility of a CFE relies on its ability to maintain temporal coherence. That is, a successful explanation must remain consistent with the underlying dynamics and realistic structure of the original sequence.

Plausibility, a crucial property of any CFE, is defined as the degree to which generated examples remain realistic and aligned with the data distribution. Existing methods for time series either enforce plausibility indirectly through decoder reconstruction constraints \cite{Wang2024,Wang2021}, rely on direct substitution of real training samples \cite{Ates2021,Delaney2021Native}, or lack explicit mechanisms to preserve realistic temporal dynamics. CFEs that introduce implausible or adversarial-looking patterns undermine user trust and fail to provide meaningful guidance \cite{guidotti2024counterfactual}. Incorporating explicit alignment with target-class samples into the optimization objective can therefore help ensure that generated CFEs remain both valid and realistic.

We address this issue by introducing a novel method for generating plausible time series CFEs through direct optimization in the input space. Plausibility is enforced via soft-DTW distance to $k$-nearest neighbors from the target class, which preserves realistic temporal patterns by aligning CFEs with actual data samples. To ensure other desired properties, such as achieving target class prediction or minimal changes, the optimization objective also involves proximity, sparsity, and validity constraints. 

Our contributions are as follows:
\begin{itemize}
\item[\textbullet] A new method for generating time series counterfactual explanations that enforces plausibility through soft-DTW alignment with target class samples while maintaining validity, proximity, and sparsity;
\item[\textbullet] A comprehensive evaluation against strong reference methods across validity, sparsity, proximity, and plausibility metrics, employing average DTW distance to
$k$-nearest neighbors as a plausibility measure;
\item[\textbullet] A qualitative analysis comparing the counterfactuals generated by different methods, illustrating the limitations of existing approaches in preserving temporal realism and demonstrating the superior plausibility of our proposed method.
\end{itemize}

\section{Related work}\label{sec:related-works}

Early CFE methods for time series rely on the direct substitution of segments from training data. CoMTE \cite{Ates2021} formulates CFE generation as an optimization problem, iteratively selecting a binary diagonal substitution matrix to replace individual time points in the test instance with corresponding values from a chosen target-class instance. In contrast, Native Guide \cite{Delaney2021Native} retrieves a nearest unlike neighbor from the training set and adapts it toward the target class by selectively replacing contiguous subsequences of the original query. This replacement is often guided by feature-weight vectors extracted from the classifier to ensure sparsity and semantic meaning. Similarly, the shapelet-based method \cite{guidotti2024counterfactual} explicitly uses interpretable features, constructing CFEs through a two-step process: first, removing original-class shapelets using a nearest unlike neighbor, and then inserting target-class shapelets. While these approaches inherit plausibility from real training samples, they constrain counterfactuals to existing data patterns and may introduce abrupt, discontinuous modifications.

Another line of work employs evolutionary algorithms for counterfactual search. TSEvo \cite{Hllig2022} applies genetic algorithms to perturb time series but does not provide explicit plausibility guarantees. Sub-SpaCE \cite{Refoyo2023} combines evolutionary search with an autoencoder trained on the original data, optimizing a multi-objective function that balances validity, sparsity, modification contiguity, and reconstruction error to encourage counterfactuals that remain within the data manifold.

Several methods incorporate deep learning models to ensure plausibility through learned representations. LatentCF++~\cite{Wang2021} and Glacier~\cite{Wang2024} utilize autoencoders to perturb time series within the latent space, optimizing validity and proximity while enforcing plausibility through reconstruction constraints. Time-CF \cite{Huang2024ShapGAN} employs a TimeGAN trained on non-target class instances to generate synthetic time series, which are then modified by identifying discriminative shapelet locations in the original series and replacing those subsequences with corresponding segments from the TimeGAN-generated samples. CELS \cite{li2023cels} and M-CELS \cite{li2024mcels} learn differentiable saliency maps to identify the most influential time steps for classification decisions. This map is then used to guide sparse perturbations (by replacing values with those from the nearest unlike neighbor) via an optimization that balances validity, sparsity, and temporal consistency. Reinforcement learning \cite{Sun2024NoDtrainMA} has also been applied to train an agent that modifies time series to achieve desired classifications, where plausibility is defined as respecting causal properties in the data rather than distributional similarity. 

More recently, MASCOTS \cite{Pudowski2025} introduces a novel, model-agnostic approach by transforming the time series into a symbolic feature space using the Bag of Receptive Fields (BoRF) representation \cite{Spinnato2024FastIA}. Operating in this learned feature space, MASCOTS generates CFEs through an iterative pattern-swapping process, identifying segments most relevant to the original class and perturbing them based on target-class symbolic patterns. This ensures that the generated explanations are highly interpretable and can be expressed visually or in natural language.

Unlike substitution-based methods that restrict CFEs to patterns observed in the training data, or autoencoder-based methods that enforce plausibility only indirectly through reconstruction, our approach explicitly aligns CFEs with target-class samples using a differentiable soft-DTW distance, enabling smooth gradient-based optimization that preserves realistic temporal dynamics.

\section{Preliminaries}
In this section, we provide the required notation and concepts used throughout this work.

\paragraph{Time series classification}
A time series is a sequence of observations $X = (x_1, x_2, \ldots, x_T) \in \mathbb{R}^{T \times d}$, where $T$ is the length of the series and $d$ is the number of channels (dimensions). A time series classifier is a function $f: \mathbb{R}^{T \times d} \rightarrow \mathcal{Y}$, where $\mathcal{Y} = \{1, 2, \ldots, c\}$ represents the set of $c$ possible class labels. Given a training dataset $\mathcal{D} = \{(X_i, y_i)\}_{i=1}^n$, the classifier $f$ is trained to predict the class label $\hat{y} = f(X)$ for an input time series $X$.

\paragraph{Counterfactual explanations}
Following the definition in \cite{guidotti2024counterfactual}, a counterfactual explanation for a time series $X$ classified as $\hat{y} = f(X)$ is a modified time series $X'$ such that: (1) $f(X') \neq \hat{y}$ (validity), (2) $X'$ is similar to $X$ (proximity), and (3) $X'$ is plausible within the target class distribution. Additional desirable properties include sparsity, which measures the extent of modifications between $X$ and $X'$. Proximity is typically measured by a distance metric such as Euclidean distance, while plausibility assesses whether $X'$ represents a realistic instance rather than an adversarial perturbation.

\paragraph{Dynamic Time Warping}
Dynamic Time Warping (DTW) \cite{Cuturi2017SoftDTW,Sakoe1978} is a similarity measure for time series that accounts for temporal misalignments. Unlike Euclidean distance, which requires point-to-point correspondence, DTW allows for flexible alignment of time series, making it robust to different kinds of distortions. This enables the comparison of time series with similar shapes but different temporal dynamics.

Given two time series $X = (x_1, \ldots, x_m) \in \mathbb{R}^{p \times m}$ and $Y = (y_1, \ldots, y_{m'}) \in \mathbb{R}^{p \times m'}$, DTW first computes a cost matrix $\Delta(X, Y) = [\delta(x_i, y_j)]_{ij} \in \mathbb{R}^{m \times m'}$, where $\delta(\cdot, \cdot)$ is a selected distance function (typically squared Euclidean distance). An alignment between $X$ and $Y$ is represented by a binary alignment matrix $A \in \mathcal{A}_{m,m'}$, where $\mathcal{A}_{m,m'}$ is the set of binary matrices corresponding to valid paths connecting $(1,1)$ to $(m,m')$ with only horizontal, vertical, and diagonal moves. The cost of an alignment $A$ is given by the inner product $\langle A, \Delta(X,Y) \rangle = \sum_{i,j} A_{ij} \delta(x_i, y_j)$. DTW finds the alignment with minimum cost:
\begin{equation}
\text{DTW}(X, Y) = \min_{A \in \mathcal{A}_{m,m'}} \langle A, \Delta(X, Y) \rangle.
\end{equation}
This optimization can be solved efficiently using dynamic programming with quadratic time complexity. However, DTW is non-differentiable at points where multiple optimal alignments exist.

\section{Method}

Gradient-based optimization in the input space enables iterative modification of a time series while controlling desired properties through differentiable loss terms. However, incorporating temporal alignment poses a challenge: although DTW is effective at capturing temporal similarity, its non-differentiability makes it unsuitable for gradient-based optimization.

To overcome this limitation, we employ soft-DTW \cite{Cuturi2017SoftDTW}, which replaces the hard minimum in DTW with a differentiable soft-minimum operator. The soft-minimum smoothly approximates the minimum over a set of values as:
\begin{equation*}
{\min}^\gamma\{a_1, \ldots, a_n\} = -\gamma \log \sum_{i=1}^{n} e^{\frac{-a_i}{\gamma}},
\end{equation*}
where $\gamma > 0$ is a smoothing parameter. Applying this to all possible alignment costs yields:
\begin{equation}
\text{DTW}^\gamma(X, Y) = {\min}^\gamma \{\langle A, \Delta(X, Y) \rangle : A \in \mathcal{A}_{m,m'}\}.
\end{equation}
As $\gamma \to 0^+$, soft-DTW converges to standard DTW. The soft-minimum considers all alignment matrices weighted by their costs, enabling gradient-based optimization while preserving DTW's temporal alignment properties.

Given an input series $X$ with predicted label $\hat{y} = f(X)$ and a chosen target class $y_{\text{target}} \neq \hat{y}$, we optimize a counterfactual $X'$ directly in input space by minimizing the objective:
\begin{equation}
\mathcal{L}_{\text{CF}} = \mathcal{L}_{\text{prox}} + \mathcal{L}_{\text{sparse}} + \lambda \cdot (\mathcal{L}_{\text{valid}} +\mathcal{L}_{\text{DTW}}),
\end{equation}
where $\lambda$ balances the contribution of validity and plausibility relative to proximity and sparsity.

Proximity enforces similarity between $X$ and $X'$ and is measured using the squared Euclidean distance:
\begin{equation*}
\mathcal{L}_{\text{prox}} = \frac{1}{d\,T} \|X' - X\|_2^2.
\end{equation*} 
This term maintains similarity to the original input, ensuring that the counterfactual represents a minimal perturbation.

Sparsity encourages localized modifications by penalizing the $\ell_1$ norm of the perturbation:
\begin{equation*}
\mathcal{L}_{\text{sparse}} = \frac{1}{d\,T} \|X' - X\|_1.
\end{equation*}
This promotes concentrating modifications in specific regions rather than perturbing the entire series.

The validity loss ensures that the classifier assigns $X'$ to the target class using hinge loss:    
\begin{equation*}
\mathcal{L}_{\text{valid}} = \max\left(0, \tau - p_f\left(y_{\text{target}}|X'\right)\right),
\end{equation*}
where $p_f$ is the classifier's predicted probability, and $\tau$ is a selected threshold. This term drives the optimization toward a confident classification in the target class.

The plausibility loss employs soft-DTW distance to align $X'$ with $k$-nearest neighbors $\mathcal{N}_k\left(X, y_{\text{target}}\right)$ from the target class:
\begin{equation*}
\mathcal{L}_{\text{DTW}} = \frac{1}{k} \sum_{Y \in \mathcal{N}_k(X, y_{\text{target}})} \text{DTW}^\gamma\left(X', Y\right).
\end{equation*}
By aligning with real target samples, this term ensures that $X'$ preserves realistic temporal patterns characteristic of the target class rather than producing adversarial perturbations.

The counterfactual is obtained via gradient descent on $X'$ for a fixed number of iterations, with classifier parameters held constant throughout the optimization.

\section{Experiments and Results}

In this section, we provide details on the experimental setup and present the obtained results in both quantitative and qualitative manner. All experiments have been conducted on a MacBook Pro M4 (14 cores, 48 GB of memory). Implementation details and code are publicly available and can be found at \\ \url{https://github.com/genwro-ai/soft-dtw-counterfactual-explantions}.

\subsection{Experiments setup}

\paragraph{\textbf{Datasets}} All datasets used in this work come from UCI and UEA repositories \cite{Dau2019UCR}. The chosen datasets encompass different characteristics: varying numbers of classes, channels, and timestamps. Details can be found in Table~\ref{tab:datasets}.

\begin{table}[ht]
\centering
\caption{Datasets details: number of samples ($n$), time series length ($T$), number of dimensions ($d$), number of classes ($c$), and accuracy score achieved by classifier.}
\label{tab:datasets}
\begin{tabular}{lccccc}
\toprule
Dataset & $n$ & $T$ & $d$ & $c$ & Accuracy (\%) \\
\midrule
\multicolumn{6}{l}{\textbf{Univariate Datasets}} \\
\midrule
CBF & 930 & 128 & 1 & 3 & 100.00 \\
TwoLeadECG & 1162 & 82 & 1 & 2 & 100.00 \\
GunPoint & 200 & 150 & 1 & 2 & 100.00 \\
Earthquakes & 461 & 512 & 1 & 2 & 78.26 \\
Coffee & 56 & 286 & 1 & 2 & 100.0 \\
ItalyPowerDemand & 1096 & 24 & 1 & 2 & 97.26 \\
\midrule
\multicolumn{6}{l}{\textbf{Multivariate Datasets}} \\
\midrule
Cricket & 180 & 1197 & 6 & 12 & 100.00 \\
Epilepsy & 275 & 206 & 3 & 4 & 96.36 \\
\bottomrule
\end{tabular}
\end{table}
\vspace{-0.2cm}

\paragraph{\textbf{Reference methods}}

We compare our method with two reference methods: Glacier \cite{Wang2024} and M-CELS \cite{li2024mcels}, described in Section~\ref{sec:related-works}. We use the settings recommended by the authors of the methods \footnote{Glacier: \url{https://github.com/zhendong3wang/learning-time-series-counterfactuals}, M-CELS: \url{https://github.com/Luckilyeee/M-CELS}}.

For Glacier, we utilize its “uniform” variant of weighting, which, as the authors showed, offers the best results. Due to the limitations of this method, we evaluated it only for univariate time series datasets.

\paragraph{\textbf{Classifier}}

Following Glacier \cite{Wang2024}, we employ a 1D convolutional neural network with three convolutional blocks (channels: 32, 64, 128), each consisting of convolution, batch normalization, ReLU activation, and max-pooling. The network uses a kernel size of 3, adaptive average pooling, and a final linear classification layer. Hyperparameters (dropout, learning rate, weight decay) are optimized using Optuna \cite{ozaki2025optunahub}, and training proceeds for up to 80 epochs with early stopping. The resulting classifier achieves strong accuracy across all datasets, as shown in the last column of Table~\ref{tab:datasets}.

\paragraph{\textbf{Evaluation metrics}}

We evaluate counterfactuals using five metrics. Validity ($\mathbf{Val} \uparrow$) measures the fraction of counterfactuals that successfully flip the classifier's prediction:
\begin{equation}
\text{Val}(X, X') = \frac{1}{n}\sum_{i=1}^n \mathbbm{1}[f(X_i) \neq f(X'_i)],
\end{equation}
where higher is better. Sparsity($\mathbf{L_1} \downarrow$) and Proximity ($\mathbf{L_2} \downarrow$) quantify the distance between the original and counterfactual time series using $L_1$ and $L_2$ norms, respectively:
\begin{equation}
L_p(X, X') = \frac{1}{n}\sum_{i=1}^n \|X_i - X'_i\|_p,
\end{equation}
where lower values indicate sparser modifications ($p=1$) or closer counterfactuals ($p=2$). Plausibility ($\mathbf{DTW} \downarrow$) assesses whether counterfactuals align with the target class distribution by using the average DTW distance to the 10 nearest neighbors from the target class, where lower distances indicate more plausible counterfactuals. Additionally, we report the Isolation Forest Score (\textbf{Iso Forest Score}~$\uparrow$), measuring the fraction of counterfactuals classified as nominal (non-outliers), where higher values indicate better plausibility.

\paragraph{\textbf{Hyperparameters tuning}}

Our method involves two key hyperparameters: $\lambda$, which control the balance between validity/plausibility and proximity/sparsity objectives, and $k$, the number of nearest target class neighbors used to guide the CFE generation. To determine optimal values, we conducted a hyperparameter search on three datasets: CBF, TwoLeadECG, and GunPoint, evaluating $\lambda \in \{1, 2, 5\}$ and $k \in \{5, 10, 20\}$ across validity, proximity (L2), plausibility (DTW), and throughput (samples per second).

\begin{figure}
    \centering
    \includegraphics[width=0.9\linewidth]{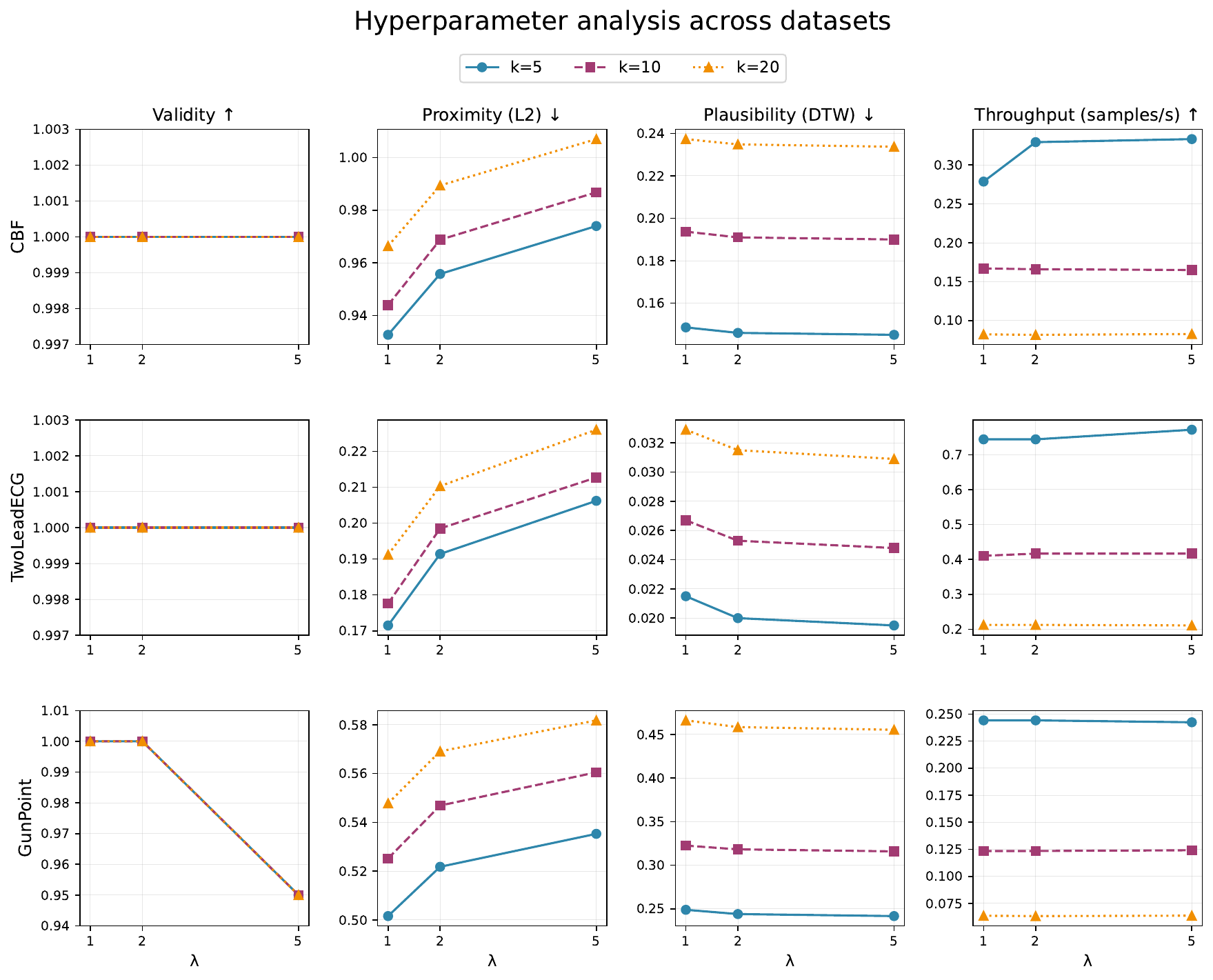}
    \caption{Analysis of hyperparametrs impact on obtained metrics and execution times across CBF, TwoLeadECG and GunPoint datasets.}
    \label{fig:hyperparams}
\end{figure}

The results, presented in Figure~\ref{fig:hyperparams}, show that while all configurations achieve high validity, proximity and plausibility degrade significantly as $k$ increases from 5 to 20. Computation efficiency decreases with larger $k$ values due to the additional DTW calculations required for alignment with more neighbors. Lower $\lambda$ values maintain better proximity and plausibility across all datasets.

Based on these results, we select $\lambda=1$ and $k=10$ as our default hyperparameters, providing a desired balance between validity, proximity, plausibility, and computational efficiency. 

Additionally, we set $\gamma = 1$ for the soft-DTW parameter, following \cite{Cuturi2017SoftDTW}, which demonstrated robust performance across multiple time series tasks.

\subsection{Quantitative results}

Table~\ref{tab:results} gathers the results for all methods across considered datasets and metrics. We analyze the performance across three key aspects: plausibility, validity, and the trade-off between proximity and sparsity.

\begin{table}[ht]
\centering
\caption{Performance comparison across different datasets and methods.}
\label{tab:results}
\begin{tabular}{llccccc}
\toprule
\textbf{Dataset} & \textbf{Method} & $\mathbf{Val} \uparrow$ &  $\mathbf{L_1} \downarrow$ & $\mathbf{L_2} \downarrow$  & $\mathbf{DTW} \downarrow$ & $\textbf{Iso Forest Score} \uparrow$ \\
\midrule
\multirow{3}{*}{\textbf{CBF}} & Ours & $\mathbf{1.000}$ & 9.871 & 1.071 & $\mathbf{0.194}$ & $\mathbf{1.000}$ \\
& Glacier & 0.360 & 4.062 & 0.540 & 1.415 & 0.987 \\
& M-Cels & 0.226 & $\mathbf{1.500}$ & $\mathbf{0.486}$ & 2.402 & 0.984 \\
\midrule
\multirow{3}{*}{\textbf{TwoLeadECG}} & Ours & $\mathbf{1.000}$ & 1.446 & 0.214 & $\mathbf{0.016}$ & $\mathbf{1.000}$ \\
& Glacier & 0.233 & 0.484 & $\mathbf{0.115}$ & 0.064 & $\mathbf{1.000}$ \\
& M-Cels & 0.970 & $\mathbf{0.245}$ & 0.119 & 0.302 & 0.879 \\
\midrule
\multirow{3}{*}{\textbf{GunPoint}} & Ours & $\mathbf{0.975}$ & $4.478$ & $0.491$ & $\mathbf{0.155}$ & $\mathbf{1.000}$ \\
& Glacier & 0.000 & 0.639 & 0.170 & 0.436  & $\mathbf{1.000}$ \\
& M-Cels & 0.425 & $\mathbf{0.129}$ & $\mathbf{0.074}$ & 2.317 & 0.925 \\
\midrule
\multirow{3}{*}{\textbf{Earthquakes}} & Ours & $\mathbf{1.000}$ & 48.985 & 2.441 & $\mathbf{0.775}$ & 0.924 \\
& Glacier & 0.000 & 8.528 & $\mathbf{0.661}$ & 1.907 & $\mathbf{1.000}$ \\
& M-Cels & 0.174 & $\mathbf{6.765}$ & 1.167 & 0.288 & $\mathbf{1.000}$ \\
\midrule
\multirow{3}{*}{\textbf{Coffee}} & Ours & $\mathbf{1.000}$ & 5.979 & 0.489 & $\mathbf{0.064}$ & $\mathbf{1.000}$ \\
& Glacier &  0.455 & 9.182 & 0.795 & 1.024 & $\mathbf{1.000}$ \\
& M-Cels & $\mathbf{1.000}$ & $\mathbf{0.527}$ & $\mathbf{0.183}$ & 0.4231 & 0.636 \\
\midrule
\multirow{3}{*}{\textbf{ItalyPowerDemand}} & Ours & $\mathbf{1.000}$ & 0.869 & 0.222 & $\mathbf{0.015}$ & $\mathbf{1.000}$ \\
& Glacier & 0.023 & 0.307 & 0.107 & 0.054 & $\mathbf{1.000}$ \\
& M-Cels & 0.466 & $\mathbf{0.178}$ & $\mathbf{0.091}$ & 0.369 & 0.831 \\
\midrule
\multirow{3}{*}{\textbf{Cricket}} & Ours & $\mathbf{1.000}$ & 475.900 & 12.210 & $\mathbf{0.810}$ & $\mathbf{0.972}$ \\
& Glacier & N/A & N/A & N/A & N/A & N/A \\
& M-Cels & 0.194 & $\mathbf{54.403}$ & $\mathbf{2.636}$ & 65.924 & 0.888 \\
\midrule
\multirow{3}{*}{\textbf{Epilepsy}} & Ours & $\mathbf{1.000}$ & 68.130 & 3.138 & $\mathbf{3.445}$ & $\mathbf{1.000}$ \\
& Glacier & N/A & N/A & N/A & N/A & N/A \\
& M-Cels & 0.272 & $\mathbf{14.807}$ & $\mathbf{1.623}$ & 19.213 & $\mathbf{1.000}$ \\
\bottomrule
\end{tabular}
\end{table}

\paragraph{\textbf{Validity}} Our method maintains perfect or near-perfect validity across all datasets, confirming its reliable ability to generate successful CFEs. In contrast, reference methods show significantly lower success rates. For instance, on the classic CBF dataset, Glacier's validity score is only 0.360, and M-CELS's is 0.226. The disparity is also visible on ItalyPowerDemand, where Glacier's score falls to a mere 0.023, and M-CELS's is 0.466.

\paragraph{\textbf{Plausibility}} Our method achieves the best plausibility scores across all datasets for the DTW distance and obtains the best or near-best Isolation Forest Scores. In case of the DTW distance, this is expected given that our optimization explicitly includes a loss component directly aligned with the DTW evaluation metric. However, the scale of improvement is notable: our method achieves DTW distances that are often an order of magnitude lower than competing approaches. For instance, on the TwoLeadECG dataset, our method obtains a DTW distance of 0.016 compared to 0.064 for Glacier and 0.302 for M-CELS. Similarly, on the Cricket dataset, our DTW distance of 0.810 contrasts sharply with M-CELS's 65.924. These substantial differences indicate that our approach generates counterfactuals that closely align with the temporal patterns of the target class. The Isolation Forest Score further supports this finding: our method achieves perfect scores (1.000) on six datasets and competitive scores on the remaining two (0.972 on Cricket and 0.924 on Earthquakes), indicating that the generated counterfactuals are recognized as nominal instances within the target class distribution rather than outliers.

\paragraph{\textbf{Proximity and Sparsity}} Our method consistently shows higher $L_1$ and $L_2$ distances compared to reference methods. For example, on the CBF dataset, our $L_1$ distance of 9.871 is larger than M-CELS's 1.500, and our $L_2$ distance of 1.071 exceeds M-CELS's 0.486. This pattern holds across all datasets. This demonstrates the trade-off between proximity and plausibility, with our method prioritizing the latter. To better understand these differences and the nature of the generated CFEs, we examine specific examples in the following section.

\subsection{Qualitative results}

\begin{figure}[ht]
    \centering
    \includegraphics[width=0.99\linewidth]{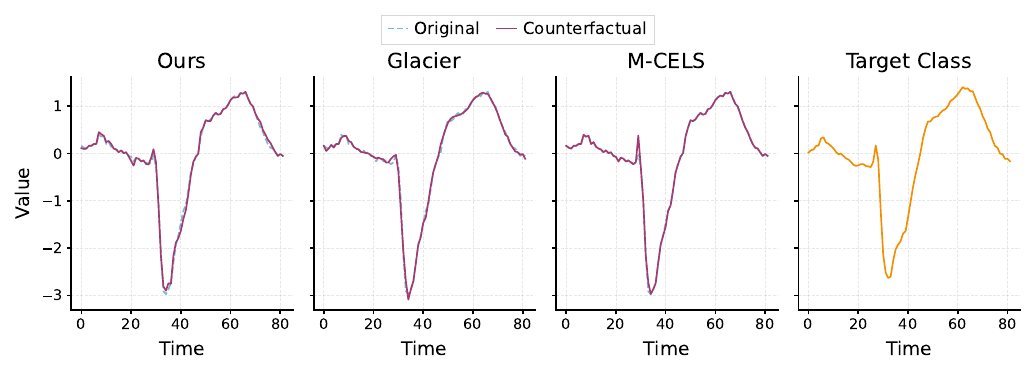}
    \caption{Exemplary counterfactual explanations generated by compared methods for TwoLeadECG dataset.}
    \label{fig:ecg-counterfactuals}
\end{figure}

\begin{figure}
    \centering
    \includegraphics[width=0.99\linewidth]{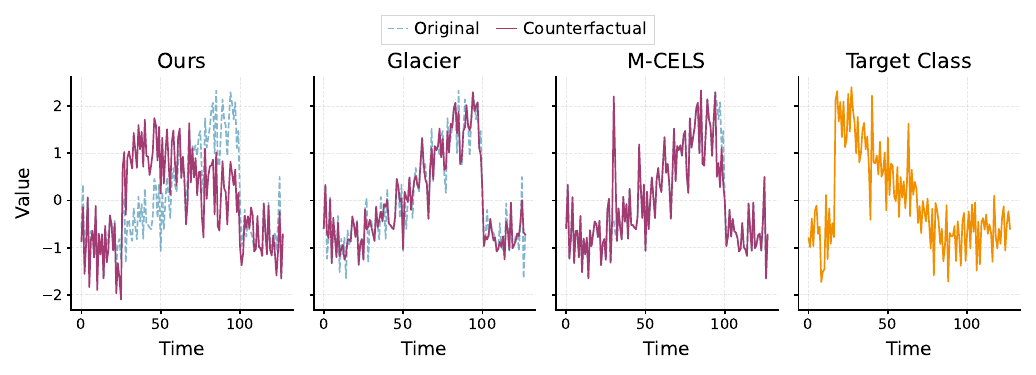}
    \caption{Exemplary counterfactual explanations generated by compared methods for CBF dataset.}
    \label{fig:cbf-counterfactuals}
\end{figure}

To illustrate the trade-off between proximity and plausibility and to better understand the structural integrity of the generated CFEs, we examine specific examples from the TwoLeadECG and CBF datasets, shown in Figures~\ref{fig:ecg-counterfactuals} and~\ref{fig:cbf-counterfactuals}, respectively. Each figure displays the original time series, the generated CFE, and representative samples from the target class.

Figure~\ref{fig:ecg-counterfactuals} presents CFEs for an ECG signal. Both our method and M-CELS successfully generate valid CFEs that closely align with the target class structure, particularly capturing the prominent peak around timestep 30. In contrast, Glacier produces more subtle modifications that deviate less from the original signal but fail to capture the prominent temporal pattern required by the target class. An interesting observation is that all three methods introduce modifications in regions where changes may not be strictly necessary for achieving validity. M-CELS demonstrates the most focused modifications, while our method and Glacier introduce more distributed perturbations throughout the sequence.

However, the differences become more apparent on the CBF dataset, as shown in Figure~\ref{fig:cbf-counterfactuals}. The CBF dataset contains three classes: Cylinder, Bell, and Funnel, each characterized by distinct geometric shapes. Our method transforms the original instance into a CFE that closely matches the temporal structure of the target class. In contrast, both Glacier and M-CELS generate CFEs that resemble adversarial attacks rather than meaningful transformations. Their modifications are minimal yet insufficient to produce CFEs that align well with the target class.

These qualitative examples illustrate the trade-off our method makes: accepting larger proximity values in favor of generating CFEs that exhibit realistic temporal patterns consistent with the target class distribution. While competing methods may achieve lower $L_1$ and $L_2$ distances in some cases, they often fail to produce valid CFEs or generate perturbations that lack temporal coherence with the target class.

\section{Limitations}
Our method has several limitations. First, the computational complexity of soft-DTW scales quadratically with sequence length \cite{Cuturi2017SoftDTW}, and computing distances to k-nearest neighbors requires additional DTW calculations at each optimization iteration. This makes the approach computationally expensive for long time series.
Second, the plausibility component relies on aligning with the $k$-nearest neighbors from the training data. This approach assumes the target class exhibits consistent temporal patterns. In reality, classes can possess complex, multi-modal distributions. When a class shows high temporal variability, $k$-nearest neighbor alignment may force the CFE to align with a pattern that does not coherently represent the target class, thus limiting the method's effectiveness in capturing high class diversity.

\section{Conclusions}

We presented a new gradient-based method for generating plausible counterfactual explanations for time series classifiers. The key contribution of our approach is the explicit enforcement of plausibility through soft-DTW alignment with $k$-nearest neighbors from the target class, which ensures the preservation of realistic temporal patterns rather than producing adversarial perturbations.

Our experimental evaluation across eight datasets demonstrates that the method consistently achieves perfect or near-perfect validity while maintaining strong plausibility scores. The soft-DTW loss component proves effective in aligning CFEs with the target class distribution, as evidenced by DTW distances that are often an order of magnitude lower than those of competing methods and high Isolation Forest Scores indicating nominal instance recognition.

The necessity of explicitly enforcing plausibility introduces an inherent trade-off between plausibility and proximity in the generated CFEs. Our results consistently show that prioritizing the preservation of realistic temporal patterns through the $\mathcal{L}_{\text{DTW}}$ term leads to an increase in the required proximity distance ($\ell_1$ and $\ell_2$ norms) when compared to methods that solely minimize perturbation. This finding confirms that achieving plausibility, particularly through alignment with real target class samples, often necessitates larger yet more meaningful modifications to the input time series.

Future work will explore the use of probabilistic generative models for modeling time series density to generate plausible CFEs. Such models could better capture the variety of temporal patterns within target classes, mitigating the limitation of k-nearest neighbor alignment. This would address scenarios where distinct temporal structures coexist within the same class, providing more coherent guidance for CFE generation.

\begin{credits}
\subsubsection{\ackname} This study was supported by the National Science Centre (Poland) Grant No. 2024/55/B/ST6/02100.

\subsubsection{\discintname}
The authors have no competing interests to declare that are relevant to the content of this article.
\end{credits}

\bibliographystyle{splncs04}
\bibliography{bib}

\end{document}